\pdfoutput=1
\documentclass{llncs}
\usepackage{multirow}
\usepackage{caption}
\usepackage{booktabs}
\usepackage{epsfig}
\usepackage{longtable}
\usepackage{rotating}
\usepackage{gensymb}
\usepackage{lscape}
\usepackage[colorlinks=true,
            linkcolor=red,
            citecolor=blue]{hyperref}
\usepackage{caption}
\usepackage{array, caption, floatrow, tabularx, makecell, booktabs}%
\begin{document}

\newcommand{\edit}[1]{\color{red}{\textbf{#1}}\color{black}{}}

\frontmatter          

\title{Differential Diagnosis for Pancreatic Cysts in CT Scans Using Densely-Connected Convolutional Networks}
\author{Hongwei Li\inst{1}, Kanru Lin\inst{3}, Maximilian Reichert\inst{2}, Lina Xu\inst{1}, Rickmer Braren\inst{2}, Deliang Fu\inst{3}, Roland Schmid\inst{2}, Ji Li\inst{3\dagger} and Bjoern Menze\inst{1*} and Kuangyu Shi\inst{2*}}
\institute{1. Department of Computer Science, Technical University of Munich\\
2. Klinikum Rechts der Isar, Technical University of Munich \\
3. Fudan University}

\maketitle              
\footnotetext{$*$ equal contribution; $\dagger$ corresponding author, email:liji@huashan.org.cn}
\begin{abstract}
The lethal nature of pancreatic ductal adenocarcinoma (PDAC) calls for early differential diagnosis of pancreatic cysts, which are identified in up to 16\% of normal subjects, and some of which may develop into PDAC.
Previous computer-aided developments have achieved certain accuracy for classification on segmented cystic lesions in CT. However, pancreatic cysts have a large variation in size and shape, and the precise segmentation of them remains rather challenging, which restricts the computer-aided interpretation of CT images acquired for differential diagnosis. We propose a computer-aided framework for early differential diagnosis of pancreatic cysts without pre-segmenting the lesions using densely-connected convolutional networks (Dense-Net). The Dense-Net learns high-level features from whole abnormal pancreas and builds mappings between medical imaging appearance to different pathological types of pancreatic cysts. To enhance the clinical applicability, we integrate saliency maps in the framework to assist the physicians to understand the decision of the deep learning method. The test on a cohort of 206 patients with 4 pathologically confirmed subtypes of pancreatic cysts has achieved an overall accuracy of 72.8\%, which is significantly higher than the baseline accuracy of 48.1\%, which strongly supports the clinical potential of our developed method.
\end{abstract}

\section{Introduction}
Pancreatic ductal adenocarcinoma (PDAC), which accounts for 90\% of malignant pancreatic cancer, has the highest mortality rates.
Despite significant advances in medicine over the past decades, it has a poor five-year survival rate of less than 5\%,  most probably because of the advanced and incurable stage of the disease at the time of diagnosis \cite{Ryan2014}. Intraductal papillary mucinous neoplasm (IPMN) and mucinous cystic neoplasm (MCN) are two precursor cystic lesions of PDAC and the early diagnosis of these precursor lesions can significantly increase the patient survival \cite{Canto2013}. However, benign cysts such as serous cystic neoplasm (SCN) and solid pseudopapillary tumor (SPT) which rarely or never give rise to malignant cancer, present very similar imaging properties as PDAC precursors. Up to 16\% of screening subjects were reported to have pancreatic cysts \cite{Reichert2011}. Unnecessary pancreas surgery of benign lesions may dramatically reduce the quality of the life of the patients. Therefore, early differential diagnosis plays a key role in such dilemma situation.

Despite of the urgent needs, there is still no clinically available method to effectively differentiate pancreatic cysts to date \cite{vincent2011pancreatic}. A recent study \cite{sahani2011prospective} reported an accuracy of 67-70\% for the discrimination of 130 pancreatic cysts on CT scans performed by two physician with 10+ years of experience in abdominal imaging.

\begin{figure*}[t]
	\begin{center}
		\includegraphics[width=0.95 \linewidth,height=0.25\linewidth]{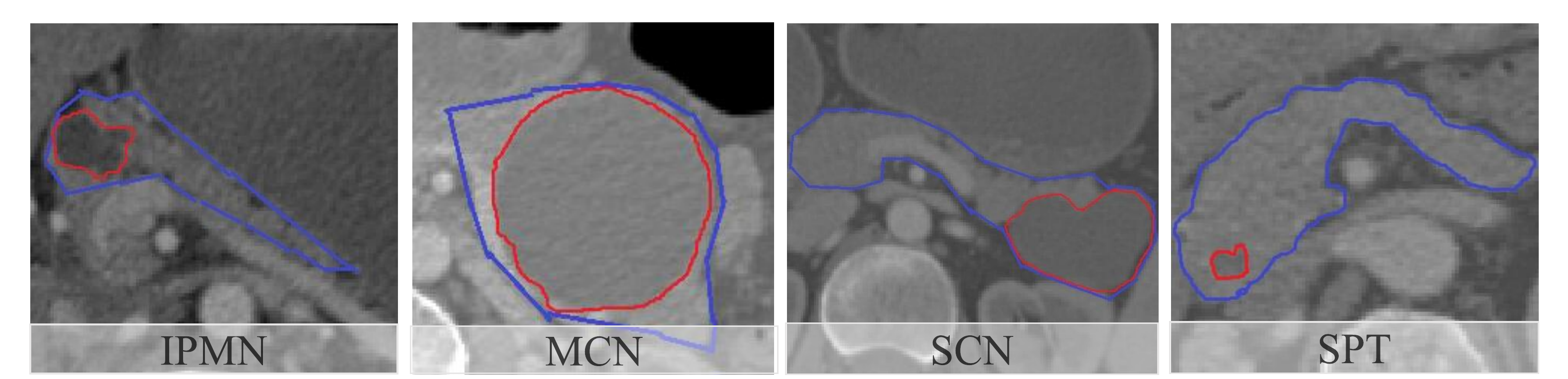}
	\end{center}
	\caption{Examples of pancreatic cyst appearance in CT images. The blue contour indicates the region of pancreas and the red contour indicates the area of cysts.}
	\label{fig:samples_cysts} 
\end{figure*}

A computer-aided diagnosis (CAD) system was proposed recently for pancreatic cysts classification \cite{dmitriev2017classification}. This method requires the cysts to be well annotated before general demographic information and texture information being extracted by convolutional neural network were aggregated by Bayesian combination. However, pancreatic cystic lesions have a large variation in size (as small as a few mm) and geometry. The heterogeneity of the lesions makes the precise identification and segmentation of the lesions extremely difficult and incur additional risks for the successful application of this method. Furthermore, focusing only on the region of cysts may lead to two technical limitations: 1) useful contextual information including the shape of abnormal pancreas caused by the inflammation is missing after masking the cysts; 2) the texture information inside some types of small cysts such as SPT shown in Fig. \ref{fig:samples_cysts} is poor.

In this paper, we present the first CAD approach for early differential diagnosis of pancreatic cysts \textbf{without the requirement of detection and segmentation of lesions} beforehand. This is achieved by using densely-connected convolutional networks (Dense-Net) \cite{huang2017densely} on whole pancreas in CT imaging, which not only learns high-level features from the whole pancreas and builds mappings from pathological types to imaging appearance, but also could generate better saliency maps to visualize important region by taking advantage of its dense connection between convolutional layers.
Four subtypes of cystic lesions, i.e.~IPMN, MCN, SCN and SPT, are classified.
To further explore contextual information from outside of the lesion, we generate saliency maps to highlight the important pixels and indicate the spatial support relevant to the decision.
This can assist the radiologists to understand the computer-aided decision, for example, to what degree the pancreas shape will make a difference in evaluating disease state.

\section{Data Preparation}
A cohort of 206 patients (81 males, 125 females, mean age 53.4 $\pm$ 15.1 years) referred to surgery because of suspected malignant pancreatic cysts were included in this study. These patients have not been reported of pancreatic disease until the diagnosis on abdominal contrast-enhanced CT scans (slice thickness 3 mm). Pathology on the surgical specimens confirmed 64 cases of IPMNs, 35 cases of MCNs, 66 cases of SCNs, and 41 cases of SPTs, which means all the benign cysts in the cohort were misdiagnosed by reading on the CT images.
Before the development of our computer-aided methods, a rough contour of the pancreas as shown in Fig. \ref{fig:samples_cysts} were defined by a junior physician. 


\section{Method}

\subsection{Densely-Connected Convolutional Networks} \label{method}

\begin{figure*}[t]
	\begin{center}
		\includegraphics[width=0.95 \linewidth,height=0.25\linewidth]{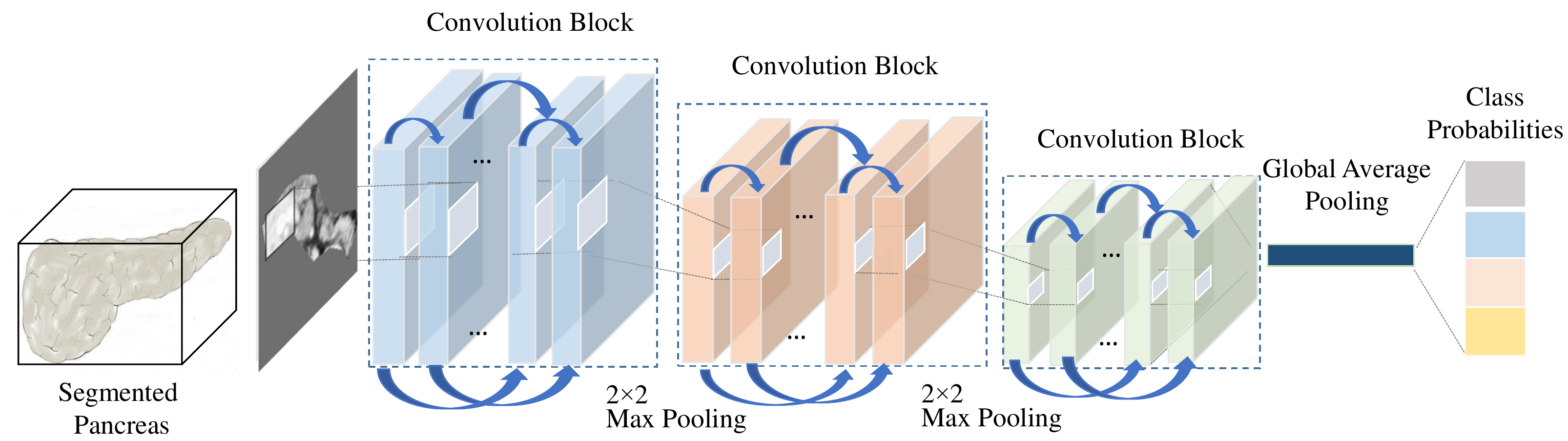}
	\end{center}
	\caption{Overview of our proposed Dense-Net architecture. It contains three convolution blocks and 10 connected convolutional layers in each block. For convolutional layers in one block, feature maps of all the previous layers are used as the inputs for each succeeding layer. 1$\times$1 convolution followed by 2$\times$2 average pooling as transition layers was employed between two contiguous dense blocks.}
	\label{fig:DenseNet} 
\end{figure*}
In this work, we explicitly learn the mapping models from imaging appearance to histopathological types via densely-connected convolutional neural networks (Dense-Net). Note that this type of CNN architecture was first proposed by \cite{huang2017densely} as a deep learning based method for general image classification. We however find that it can be a suitable model for high-level feature extraction of pancreas as well as visualization of dominant locations.
The Dense-Net tries to address two important issues: 1) dealing with the vanishing or exploding gradient problem \cite{bengio1994learning} by strengthening feature propagation; 2) reducing computation complexity by promoting feature reuse,
which strengthen its power on both image feature representation and generating gradient-based saliency maps.
In Dense-Net, each layer is connected with every other layers in the network by skip connection. Thus, the feature maps of all the previous layers are used as the inputs for each succeeding layer.
The \emph{l} layer concatenates preceding layer's feature maps, x$_{1}$, x$_{2}$, ..., x$_{l-1}$ as its input:

	\begin{equation}
	\emph{x$_{l}$} = \emph{H$_{l}$([x$_{1}$, x$_{2}$, ..., x$_{l-1}$])}
	\end{equation}

where [x$_{1}$, x$_{2}$, ..., x$_{l-1}$] represents the feature map concatenation of the ones generated in layers \emph{0}, \emph{1}, ..., \emph{l-1}. And \emph{H$_{l}$} which follows the improved residual networks \cite{he2016deep}, is a compound function of ReLUs \cite{nair2010rectified}, preceded by Batch Normalization \cite{ioffe2015batch}, and is followed by convolution.
If each function \emph{H$_{l}$} produces \emph{k} feature maps, it follows that the \emph{m$^{th}$} layer has \emph{k$_{0}$+k$\times${m}} input feature-maps, where \emph{k$_{0}$} is the number of channels in the input layer.
The hyperparameter \emph{k} is referred as \emph{growth rate} in the network.

In our classification task, the image size of the bounding box for the pancreas is 144$\times$144 and the class number is only 4, thus we choose to modify the original network designed for ImageNet in order to reduce computation complexity.
Furthermore, by reducing the number of convolution blocks, we could gain better gradients for generating the saliency maps with less risk of causing gradient vanishing.

\subsubsection{Implementation Details}
We apply Dense-Net as a high-level feature extractor and classifier, which aims to better learn shape and texture information of the pancreas.
The proposed Dense-Net, as is shown in Fig. \ref{fig:DenseNet}, contains 3 convolutional blocks, 2 max-pooling layers between each block,
1 averaged global pooling layer. For convolutional layers with kernel size 3$\times$3, each side of the inputs is zero-padded
by one pixel in order to keep the feature-map size fixed.
We use 1$\times$1 convolution followed by 2$\times$2 average pooling as transition layers between two contiguous dense blocks.
At the end of the last dense block, a global average pooling is performed and then a softmax classifier is employed.
We set number of layers \emph{L = 10} for each block, growth rate \emph{k = 9}, number of feature maps of first layer \emph{k$_{0}$ = 2k}. Bottleneck layer (1$\times$1 convolutional layer) before each 3$\times$3 convolution was used to reduce the number of input feature-maps, and thus to improve computational efficiency. For comparison with traditional CNN, we designed a similar architecture as the model in \cite{dmitriev2017classification} which specifically tailoring to pancreatic cysts classification.

\subsubsection{Training and Testing}
The data for training and testing the proposed Dense-Net were generated as follows.
Each \emph{2D} axial slice X$_{ij}^{Slice}$ of the original $3D$ bounding box \{X$_{ij}^{Slice}$\} with a
segmented pancreas x$_{i}$ was bounded to 144 $\times$ 144 squares pixels.
Due to the generally near-spherical shape of a pancreas head, slices close to the top or bottom of the volume do not contain enough pixels of a pancreas to make an accurate diagnosis. Therefore, slices with the overlap ratio of less than 10\%, defined as the
percentage of pancreas pixels in a slice, were excluded.
Overfitting of network was further reduced by applying data augmentation: 1) random rotations within a degree range of [$−$25\degree, $+$25\degree]; 2) random zoom within a range of [0.9, 1.2]; 3) random vertical flips.
The network was implemented using the Keras \cite{chollet2015keras} library and trained on 40-sized mini-batches to minimize the class-balanced cross-entropy loss function using Stochastic Gradient Descent with a 0.0005 learning rate for 100 epochs.
In the testing phase, each slice with the overlap ratio of more than 10\% was analyzed by the Dense-Net separately, and the final
probabilities were obtained by averaging the class probabilities of each slice:
	\begin{equation}
	\widetilde{P}_{DenseNet}(y_{m} = y|\{X_{ij}^{Slice}\}) = \frac{1}{N_{m}}\sum_{j=1}^{N_{m}}P_{DenseNet}(y_{m} = y|X_{mj}^{Slice})
	\end{equation}

where P$_{DensetNet}$(y$_{m}$ = y$|$X$_{mj}^{Slice}$) is the vector of class probabilities, and N$_{m}$ is the number of 2D axial slices used for the classification of pancreas sample x$_{m}$.

\subsection{Saliency Maps for Computer-Aided Analysis} \label{section_saliency_maps}
The saliency maps which visualize the dominant locations are employed as a computer-aided analysis tool in our study. We generated the saliency maps of testing images by using Guided Back-propagation method in \cite{springenberg2014striving}. It visualized the part of an input image that mostly activates a given neuron and used a simple backward pass of the activation of a single neuron after a forward pass through the network. To this end, it computed the gradient of the activation w.r.t. the image. These gradients were then used to highlight input regions that cause the most changes to the output. In each convolutional block of Dense-Net, every convolutional layer is connected to others, which address the vanishing gradient problem, thus we can gain better gradients for generating the saliency maps.

\section{Results and Discussion}
We evaluated the performance of the proposed method using a stratified 10-fold cross-validation strategy, maintaining similar data distribution in training and testing datasets considering the imbalance in the dataset. For each cross-validation, we trained a DenseNet on training set and evaluated on test set. Classification performance is reported in terms of the normalized averaged confusion matrix and the overall classification accuracy.

\subsubsection{Results of Dense-Net}

\begin{table*}
\caption{Confusion matrices of the Dense-Net (left) and traditional CNN (right). The figures in red indicate better accuracy achieved by \emph{Dense-Net} than traditional CNN while the ones in bold indicate the accuracies of each class.}\label{table:Results}.
\begin{tabular}{c}
    \begin{minipage}{0.51\linewidth}

        \begin{tabular}{ c | c | c | c | c  }
		\hline
		\textbf{Types}&~~IPMN~~&~~MCN~~&~~SCN~~&~~SPT\\
		\hline
		IPMN& \color{red}\textbf{81.25}\% & 0.03\%  & 17.19$\%$ & 3.13$\%$\\
        {MCN}& 11.43\% & \textbf{65.71}\%  &8.57 $\%$ & 14.29$\%$\\
		{SCN}& 18.18\% & {1.51}\% &\color{red}\textbf{75.76}$\%$ & 4.54$\%$\\
        {SPT}& 12.20\% & {2.43}\% & 24.39 $\%$ & \color{red}\textbf{60.98}$\%$\\
        \hline
        \end{tabular}
    \end{minipage}
\bigskip
    \begin{minipage}{.51\linewidth}
        \begin{tabular}{ c | c | c | c |c }
		\hline
		\textbf{Types}&~~IPMN~~&~~MCN~~&~~SCN~~&~~SPT\\
		\hline
		IPMN& \textbf{67.19}\% & 4.69\%  & 23.44$\%$ & 3.13$\%$\\
        {MCN}& 17.14\% & \textbf{65.71}\%  &11.43 $\%$ & 5.71$\%$\\
		{SCN}& 21.21\% & {3.03}\% & \textbf{66.67}$\%$ & 9.09$\%$\\
        {SPT}& 12.20\% & {2.43}\% & 39.02 $\%$ & \textbf{48.78}$\%$\\
        \hline
        \end{tabular}
    \end{minipage}
\end{tabular}

\end{table*}

The proposed method achieved an overall accuracy of \emph{72.8\%}, which is significantly higher than the baseline diagnostic accuracy of \emph{48.1\%} by manual reading.
From Table \ref{table:Results}, we observed that SCNs were easily misclassified into IPMNs, which is consistent with the diagnostic experiences of physicians on this cohort.
By further looking into the misclassified cases, we found that the misclassified cases of SCNs have a very similar appearance to the IPMNs while some pancreas with SCNs have a very different shape with that of other histological types.
In addition, we evaluated the benefit of using data augmention and found an overall benefit of 10.7\% as an average over all cases and classes. 

To assess the performance of the proposed Dense-Net, the results were compared with conventional CNN (see Table \ref{table:Results}), which is widely employed in computer-aided diagnosis. A same training strategy was employed for the two models during the training process.
We found that the Dense-Net outperforms traditional CNN on three classes.
Interestingly, in the MCN type, although Dense-Net and traditional CNN achieved same accuracy, they have different distributions on the misclassified types.
This could be explained as that Dense-Net distinguishes from traditional CNN in architectural  realization, especially regarding to the feature connection schema, which might be the direct cause of varied feature representations for organs.


\subsubsection{Results of Saliency Maps}
\begin{figure*}[t]
	\begin{center}
		\includegraphics[width=0.95 \linewidth,height=0.70\linewidth]{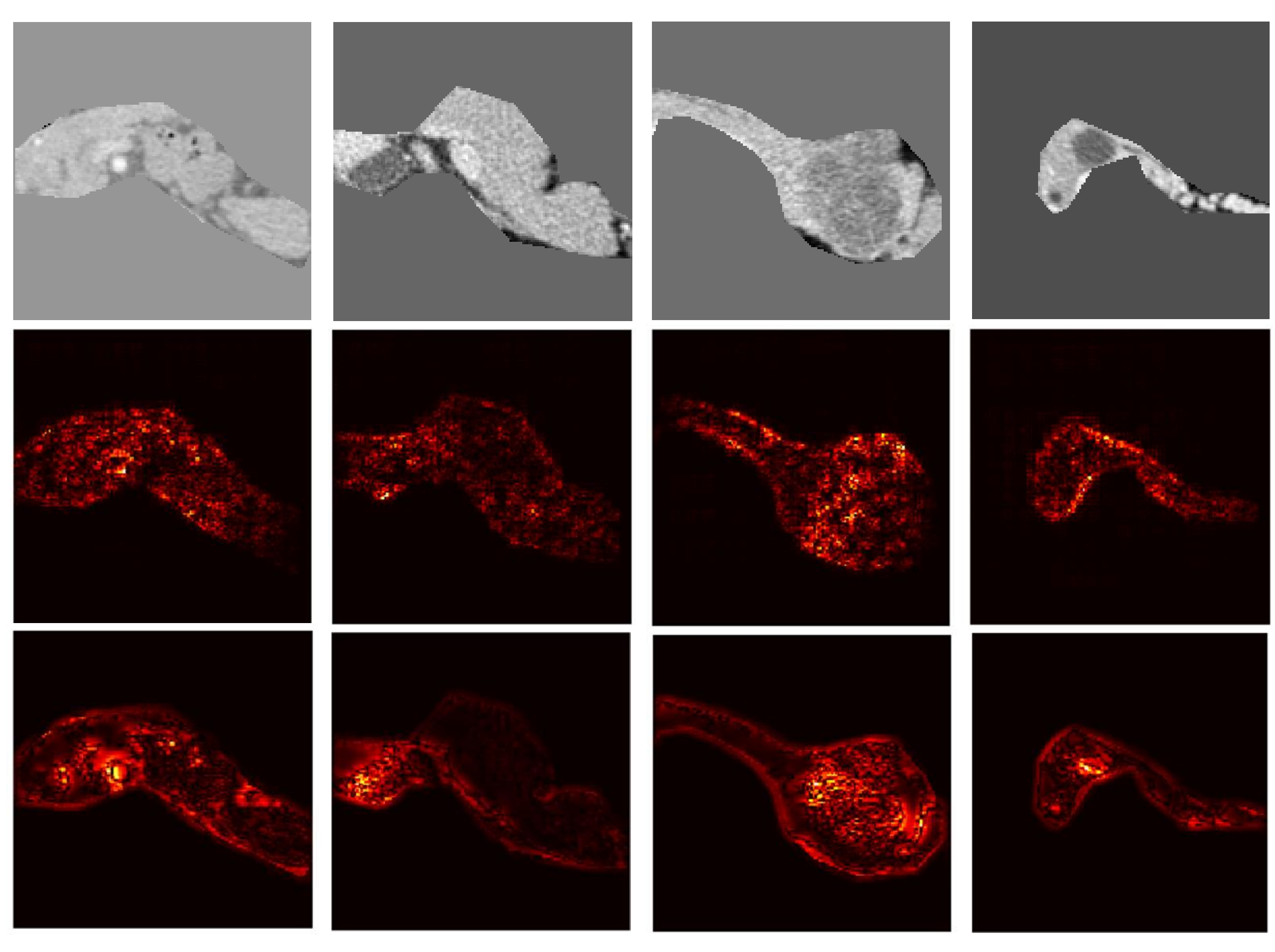}
	\end{center}
	\caption{Samples of saliency maps. From top to down: original images and saliency maps generated by traditional CNN and Dense-Net respectively. From left to right: one of the axial slices of IPMN, MCN, SCN and SPT respectively. Compared to traditional CNN, Dense-Net gains better visualization because its uniqueness on connection between convolutional layers.}
	\label{fig:saliency_maps} 
\end{figure*}

We visualized the testing images by using the gradient-based method introduced in Section \ref{section_saliency_maps},
to show the critical regions that contributed to the classification results.
Some results were shown in Fig. \ref{fig:saliency_maps}.
From the 3$^{rd}$ row, we observed that the pancreas cysts were highlighted on the four cases, demonstrating that Dense-Net extracted rich information from the region of cysts. As a comparison, the maps generated via traditional CNN were presented in 2$^{nd}$ row, which did not highlight pixels related to abnormal region. There are two main reasons: 1) the Dense-Net is more powerful than traditional CNN in feature extraction and it focuses on both abnormal region as well as shape information; 2) the connections between each two convolutional layers guarantee that Dense-Net gains better gradients for visualization than traditional CNN.
For most cases, the boundaries of pancreas were highlighted, it turns out that the shape information of pancreas can also contribute to the decision, in particular around a large lesion that has modified the organ.

\section{Summary and Conclusion}
In this work, we proposed a computer-aided diagnosis and visualization system to identify and classify pancreatic cyst.
The proposed algorithm is based on densely-connected convolutional networks in order to utilize fine imaging information from CT scans and highlight the dominant locations/pixels which not only justify the effectiveness of our model but also could serve as computer-aided diagnosis tool in clinical practice. Although the reported statistics are not attractive at first glance, it has improved 51.4\% of accuracy compared to the baseline manual diagnosis. In fact, the relatively low accuracy of many reports on differential diagnosis of pancreatic cysts were based on similar challenging data, which contained a large number of misdiagnosed cases. However, differential diagnosis on pancreatic cyst is extremely difficult. Even fine needle biopsy suffer from the heterogeneity of lesions and the inaccurate sampling due to pancreas motion. Only pathology on surgical specimens is the gold standard at the moment, which restricts the recruitment of patient cohorts. Therefore, a small improvement of the accuracy may already contribute to the state-of-the-art. The significantly improved accuracy and easy application strongly support the clinical potential of our developed method.
A limitation of our method is the need of a rough segmentation of pancreas before applying the computer-aided approach. Nevertheless, compared to the detection and segmentation of cystic lesions, the approximate segmentation of this organ avoids demanding performance burdens. Furthermore, there exists various work focusing on segmentation of whole pancreas \cite{roth2015deep,cai2017pancreas} and their results are good enough as an initial step for our method. Further improvement of the accuracy of our method can be achieved by including demographic information for classification~\cite{dmitriev2017classification}, which will be integrated as future work. For example, the incidence of SPT typically afflicts young women. The inclusion of gender and age may largely improve the accuracy for SPT.


{
\bibliographystyle{splncs03}
\bibliography{egbib}

\begin{thebibliography}{10}
\providecommand{\url}[1]{\texttt{#1}}
\providecommand{\urlprefix}{URL }

\bibitem{bengio1994learning}
Bengio, Y., Simard, P., Frasconi, P.: Learning long-term dependencies with
  gradient descent is difficult. IEEE transactions on neural networks  5(2),
  157--166 (1994)

\bibitem{cai2017pancreas}
Cai, J., Lu, L., Xie, Y., Xing, F., Yang, L.: Pancreas segmentation in mri
  using graph-based decision fusion on convolutional neural networks. In:
  {MICCAI} 2017. pp. 674--682 (2017)

\bibitem{Canto2013}
Canto, M.I., Harinck, F., Hruban, R.H., Offerhaus, G.J., Poley, J.W., Kamel,
  I., Nio, Y., Schulick, R.S., Bassi, C., Kluijt, I., Levy, M.J., Chak, A.,
  Fockens, P., Goggins, M., Bruno, M.: International cancer of the pancreas
  screening (caps) consortium summit on the management of patients with
  increased risk for familial pancreatic cancer. Gut  62(3),  339--47 (2013)

\bibitem{chollet2015keras}
Chollet, F., et~al.: Keras. \url{https://github.com/fchollet/keras} (2015)

\bibitem{dmitriev2017classification}
Dmitriev, K., Kaufman, A.E., Javed, A.A., Hruban, R.H., Fishman, E.K., Lennon,
  A.M., Saltz, J.H.: Classification of pancreatic cysts in computed tomography
  images using a random forest and convolutional neural network ensemble. In:
  {MICCAI} 2017. pp. 150--158 (2017)

\bibitem{he2016deep}
He, K., Zhang, X., Ren, S., Sun, J.: Deep residual learning for image
  recognition. In: Proceedings of the IEEE conference on computer vision and
  pattern recognition. pp. 770--778 (2016)

\bibitem{huang2017densely}
Huang, G., Liu, Z., Weinberger, K.Q., van~der Maaten, L.: Densely connected
  convolutional networks. In: {CVPR} 2017. vol.~1, p.~3 (2017)

\bibitem{ioffe2015batch}
Ioffe, S., Szegedy, C.: Batch normalization: Accelerating deep network training
  by reducing internal covariate shift. In: {ICML} 2015. pp. 448--456 (2015)

\bibitem{nair2010rectified}
Nair, V., Hinton, G.E.: Rectified linear units improve restricted boltzmann
  machines. In: {ICML} 2010. pp. 807--814 (2010)

\bibitem{Reichert2011}
Reichert, M., Rustgi, A.K.: Pancreatic ductal cells in development,
  regeneration, and neoplasia. J Clin Invest  121(12),  4572--8 (2011)

\bibitem{roth2015deep}
Roth, H.R., Farag, A., Lu, L., Turkbey, E.B., Summers, R.M.: Deep convolutional
  networks for pancreas segmentation in ct imaging. In: Medical Imaging 2015:
  Image Processing. vol. 9413, p. 94131G (2015)

\bibitem{Ryan2014}
Ryan, D.P., Hong, T.S., Bardeesy, N.: Pancreatic adenocarcinoma. N Engl J Med
  371(11),  1039--49 (2014)

\bibitem{sahani2011prospective}
Sahani, D.V., Sainani, N.I., Blake, M.A., Crippa, S., Mino-Kenudson, M., del
  Castillo, C.F.: Prospective evaluation of reader performance on mdct in
  characterization of cystic pancreatic lesions and prediction of cyst biologic
  aggressiveness. American Journal of Roentgenology  197(1),  W53--W61 (2011)

\bibitem{springenberg2014striving}
Springenberg, J.T., Dosovitskiy, A., Brox, T., Riedmiller, M.: Striving for
  simplicity: The all convolutional net. arXiv preprint arXiv:1412.6806  (2014)

\bibitem{vincent2011pancreatic}
Vincent, A., Herman, J., Schulick, R., Hruban, R.H., Goggins, M.: Pancreatic
  cancer. The Lancet  378(9791),  607--620 (2011)

\end{thebibliography}
}

\end{document}